\documentclass[10pt,twocolumn,letterpaper]{article}

\usepackage{cvpr}
\usepackage{times}
\usepackage{epsfig}
\usepackage{graphicx}
\usepackage{amsmath}
\usepackage{amssymb}
\usepackage{booktabs} 
\usepackage{algorithm}
\usepackage{algorithmic}
\usepackage{amsthm}
\usepackage{diagbox}
\usepackage{subfigure}
\usepackage{caption}
\usepackage{float}
\usepackage{epstopdf}
\usepackage{enumitem}

\newtheorem{theorem}{Theorem}

\newtheorem{remark}{Remark}

\newtheorem{definition}{Definition}

\usepackage[breaklinks=true,bookmarks=false]{hyperref}

\cvprfinalcopy 

\def\cvprPaperID{208} 

\begin{document}

\title{Progressive Feature Alignment for Unsupervised Domain Adaptation}

\author{Chaoqi Chen$^{*1}$, Weiping Xie\thanks{~indicates equal contributions.}\;$^{1}$, Wenbing Huang$^2$, Yu Rong$^2$, Xinghao Ding$^1$,\\
 Yue Huang$^{\dag1}$, Tingyang Xu\thanks{Corresponding authors}\;$^{2}$, Junzhou Huang$^2$\\
 {$^1$~Fujian Key Laboratory of Sensing and Computing for Smart City,}\\
 {School of Information Science and Engineering, Xiamen University, China}\\
 {$^2$~Tencent AI Lab}\\
 {\tt\small cqchen94@stu.xmu.edu.cn, xiewp@stu.xmu.edu.cn, hwenbing@126.com, yu.rong@hotmail.com}\\
 {\tt\small dxh@xmu.edu.cn, huangyue05@gmail.com, Tingyangxu@tencent.com, jzhuang@uta.edu}
}

\maketitle
\pagestyle{empty}
\thispagestyle{empty}

\begin{abstract}
Unsupervised domain adaptation (UDA) transfers knowledge from a label-rich source domain to a fully-unlabeled target domain. To tackle this task, recent approaches resort to discriminative domain transfer in virtue of  pseudo-labels to enforce the class-level distribution alignment across the source and target domains. These methods, however, are vulnerable to the error accumulation and thus incapable of preserving cross-domain category consistency, as the pseudo-labeling accuracy is not guaranteed explicitly. In this paper, we propose the Progressive Feature Alignment Network (PFAN) to align the discriminative features across domains progressively and effectively, via exploiting the intra-class variation in the target domain. To be specific, we first develop an Easy-to-Hard Transfer Strategy (EHTS) and an Adaptive Prototype Alignment (APA) step to train our model iteratively and alternatively. Moreover, upon observing that a good domain adaptation usually requires a non-saturated source classifier, we consider a simple yet efficient way to retard the convergence speed of the source classification loss by further involving a temperature variate into the soft-max function. The extensive experimental results reveal that the proposed PFAN exceeds the state-of-the-art performance on three UDA datasets.
\end{abstract}
\vspace{-0.5cm}
\section{Introduction}
\label{introduction}
Hiving large-scale labeled datasets is one of the reasons for the recent success of deep convolutional neural networks (CNNs) \cite{he2016deep}. Nevertheless, the collection and annotation of numerous samples in various domains is an extremely expensive and time-consuming process. Meanwhile, traditional CNNs trained on one large dataset show low generalization ability on another due to the data bias or shift \cite{torralba2011unbiased}.
\begin{figure}[!t]
\centering
\setlength{\belowcaptionskip}{-0.5cm}
\includegraphics[width=7cm]{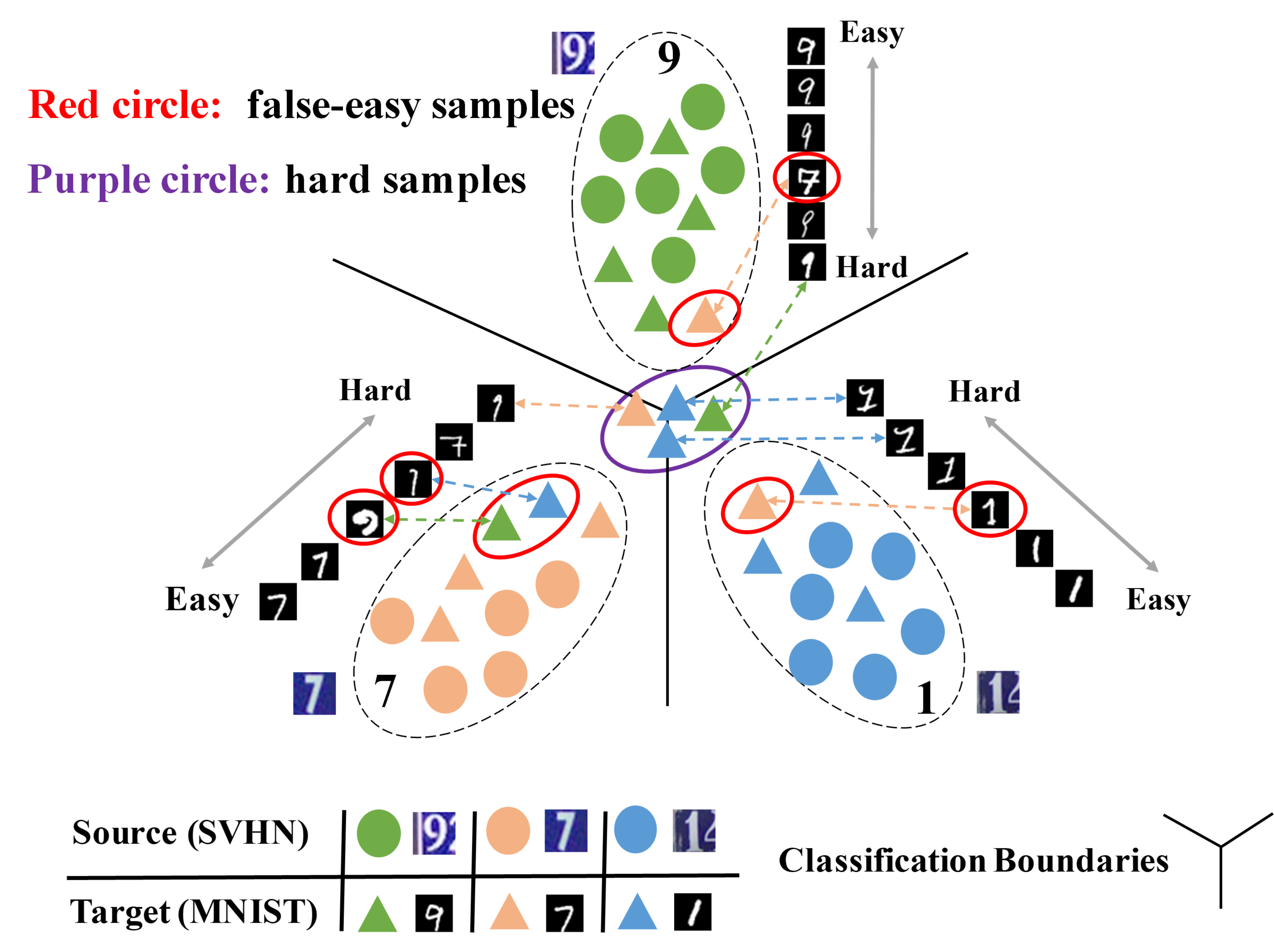}
\caption{ (Best viewed in color.) Motivations of the proposed work (SVHN$\rightarrow$MNIST). The classification boundaries are first drawn by the fully labeled source domain. There is intra-class variation in the target domain.
}\label{fig1}
\end{figure}
Unsupervised domain adaptation (UDA) methods tackle the mentioned problem by transferring knowledge from a label-rich source domain to a fully unlabeled target domain \cite{pan2010survey,pan2011domain}. 
The deep UDA methods have achieved remarkable performance \cite{tzeng2014deep,long2015learning,ganin2015unsupervised,gong2016domain,bousmalis2016domain,tzeng2017adversarial,long2017deep,saito2017maximum,pinheiro2018unsupervised,hoffman2017cycada}, which usually seek to jointly achieve small source generalization error and cross-domain distribution discrepancy.

Most prior efforts focus on matching global source and target data distributions to learn domain-invariant representations. However, the learned representations may not only bring the source and target domains closer, but also mix samples with different class labels together. Recent studies \cite{long2013transfer,sener2016learning,haeusser2017associative,saito2017asymmetric,zhang2018collaborative,pei2018multi,li2018domain,shu2018dirt,zhang2018collaborative,xie2018learning} started to consider learning discriminative representations for the target domain. Specifically, some of them \cite{sener2016learning,saito2017asymmetric,zhang2018collaborative} proposed to use pseudo-labels to learn target discriminative representations, which encourages a low-density separation between classes in the target domain \cite{lee2013pseudo}. 
Despite their efficacy, these approaches faces two critical limitations. Firstly, they require a strong pre-assumption that the correctly-pseudo-labeled samples can reduce the bias caused by falsely-pseudo-labeled samples. Nevertheless, it is challenged to satisfy the assumption, especially when the domain discrepancy is large.
The learned classifiers might be incapable of confidently distinguishing target samples, or precisely pseudo-label them with an expected accuracy requirement.
Secondly, they backpropagate the category loss for target samples based on pseudo-labeled samples, which makes the target performance vulnerable to the error accumulation.

During the exploration, we empirically observe the distinct data patterns in the target domain. The motivation is demonstrated in Fig.~\ref{fig1}. The intra-class distribution variance exists in the target domain. Some target samples, which we call easy samples, are very likely to be classified correctly since they are sufficiently close to the source domain, and we can directly assign pseudo-labels to them without any adaptation. Some target samples, which we call hard samples, lay far away from the source domain and they are ambiguous for the classification boundaries. Moreover, some easy samples, which we call false-easy samples, lay in the support of non-corresponding source classes and 
are prone to be falsely pseudo-labeled with high confidence.
These false-labeled samples introduce wrong information in the category alignment and potentially result in the error accumulation. Thus it is prerequisite to alleviate their negative influences in the context of UDA.

In this paper, we propose a Progressive Feature Alignment Network (PFAN), which largely extends the ability of prior discriminative representations-based approaches by explicitly enforcing the category alignment in a progressive manner. Firstly, an Easy-to-Hard Transfer Strategy (EHTS) progressively selects reliable pseudo-labeled target samples with cross-domain similarity measurements. However, the selected samples may include some misclassified target samples with high confidence. Then, to suppress the negative influence of falsely-labeled samples, we propose an Adaptive Prototype Alignment (APA) to align the source and target prototypes for each category. Rather than backpropagating the category loss for target samples based on pseudo-labeled samples, our work statistically align the cross-domain class distributions based on the source samples and the selected pseudo-labeled target samples. 

The EHTS and APA update iteratively and alternatively, where EHTS boosts the robustness of APA by providing reliable pseudo-labeled samples, and the cross-domain category alignment learned by APA can effectively alleviate those falsely-labeled samples introduced by the EHTS. Moreover, upon observing that a good adaptation model usually requires a non-saturated source classifier, we consider a simple yet efficient way to retard the convergence speed of the source classification loss by further involving a temperature variate into the soft-max function.
The experimental results reveal that the proposed PFAN exceeds the state-of-the-art performance on three UDA datasets.
\vspace{-0.3cm}
\section{Related Work}
We summarize the work most relevant to our proposed approach. We focus primarily on deep UDA methods due to their empirical superiority in this problem.

Inspired by the recent success of generative adversarial networks (GAN) \cite{goodfellow2014generative}, deep adversarial domain adaptation has received increasing attention in learning domain-invariant representations to reduce the domain discrepancy and provide remarkable results \cite{ganin2015unsupervised,tzeng2017adversarial,pei2018multi,zhang2018importance,zhang2018collaborative,kang2018deep,zou2018unsupervised}. These methods try to find a feature space such that confusion between the source and the target distributions in that space is maximal. For example, \cite{ganin2015unsupervised} proposed a gradient reversal layer to train a feature extractor that produces features which maximize the domain binary classifier loss, while simultaneously minimizing the label predictor loss.

Many approaches utilize a distance metric to measure the domain discrepancy between the source and target domains, such as maximum mean discrepancy (MMD), KL-divergence or Wasserstein distance \cite{gretton2012kernel,long2015learning,sun2016deep,long2016unsupervised,yan2017mind,chen2018re}. Most prior efforts intend to achieve domain alignment by matching $P(X_s)$ and $P(X_t)$. However, an exact domain-level alignment does not imply a fine-grained class-to-class overlap. Thus, it is important to pursue the category-level alignment under the absence of target true labels.

\cite{bruzzone2010domain,chen2011co,long2013transfer,sener2016learning,saito2017asymmetric,zhang2018collaborative,xie2018learning} utilize the pseudo-labels to compensate the lack of categorical information in the target domain. 
\cite{long2013transfer} jointly matched both the marginal distribution and conditional distribution using a revised MMD.
\cite{saito2017asymmetric} utilized an asymmetric tri-training strategy to learn discriminative representations for the target domain. \cite{zhang2018collaborative} iteratively selected pseudo-labeled target samples based on the classifier from the previous training epoch and re-trained the model by using the enlarged training set. \cite{xie2018learning} proposed to assign pseudo-labels to all target samples and utilize them to achieve semantic alignment across domains. However, these approaches highly relied on the hypothesis that correctly-pseudo-labeled samples can reduce the bias caused by falsely-pseudo-labeled samples. They do not explicitly alleviate those falsely-pseudo-labeled samples. When the falsely-pseudo-labeled samples take the prominent position, their performances will be limited. 
\begin{figure*}[!t]
\centering
\setlength{\belowcaptionskip}{-0.2cm}
\includegraphics[width=15.5cm]{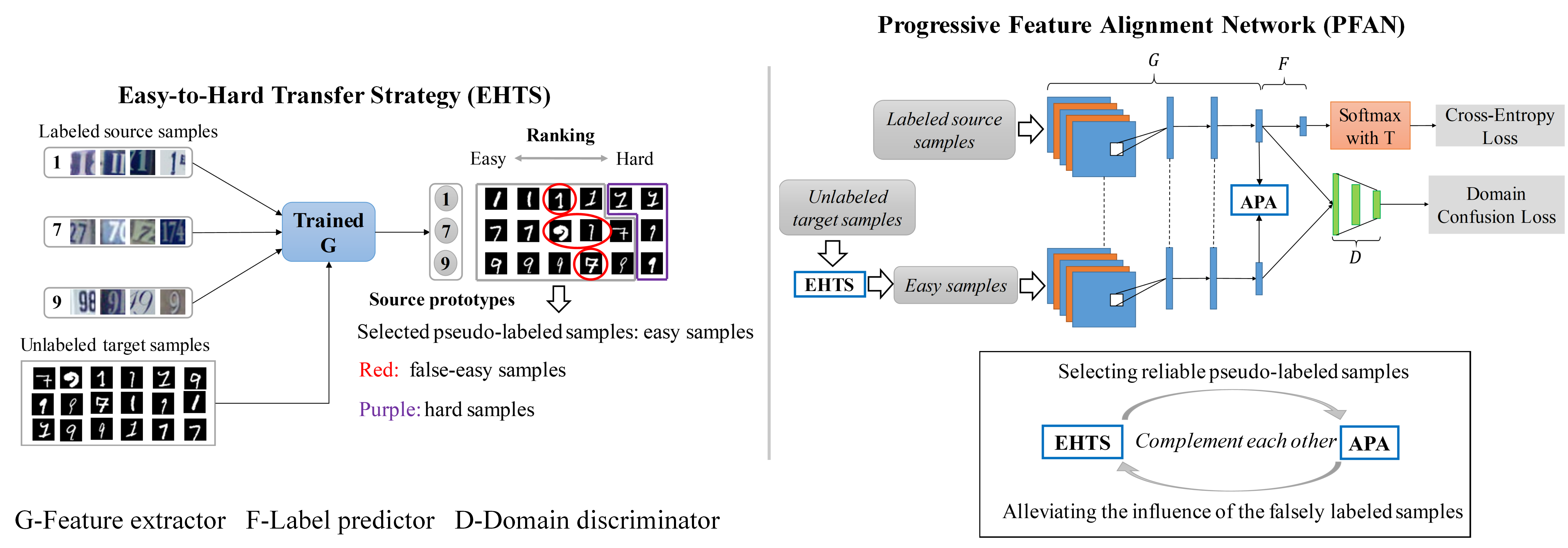}
\caption{The overall structure of the proposed PFAN. We separate the network into three modules: feature extractor $G$, label predictor $F$, domain discriminator $D$ and with associated parameters $\theta_g, \theta_{f}, \theta_d$. \textbf{Left}: The easy-to-hard strategy (ETHS). \textbf{Right}: The network structure: the dotted lines in PFAN denote weight-sharing. 
\vspace{-0.2cm}
}\label{fig2}
\end{figure*}
\vspace{-0.3cm}
\section{Progressive Feature Alignment Network}
In this section, we first provide the details of the proposed PFAN and then theoretically investigate the effectiveness of our approach. The overall architecture of PFAN is depicted in Fig.~\ref{fig2}, which consists of three components, EHTS, APA, and the soft-max function with  a temperature variate. EHTS provides reliable pseudo-labeld samples from easy to hard by iterations and APA explicitly enforces the cross-domain category alignment.
\subsection{Task Formulation}
In UDA, we are given a source domain $D_s=\{({x_i^s},{y_i^s})\}_{i=1}^{n_s}$ (${x_i^s}\in{X_s}$, ${y_i^s}\in{Y_s}$) of $n_s$ labeled samples and given a target domain $D_t=\{{x_j^t}\}_{j=1}^{n_t}$ (${x_j}^t\in{X_t}$) of $n_t$ unlabeled samples~\cite{pan2010survey}. The source and target domains are drawn from the joint probability distributions $P(X_s,Y_s)$ and $Q(X_t,Y_t)$ respectively, and ${P}\neq{Q}$. We assume that the source and target domains contain the same object classes, and we consider $C$ classes in all. 

\subsection{Easy-to-Hard Transfer Strategy}
The EHTS is biased to favor easier samples and this bias helps to avoid including the hard samples which are more likely to be given false pseudo-labels. In our approach, the easy samples are increasing progressively. Thus the ``hard'' samples will potentially be selected in further steps. The selected pseudo-labeled samples by EHTS can be used to align with their corresponding source categories as described in Section~\ref{sec:APA}.

The EHTS first computes a $D$-dimensional prototype $c_k^\mathcal{S}\in{R^D}$ of each class in the source domain. The source prototype is a mean vector of the embedded source samples in each class through an embedding function $G$ (\emph{i.e.} the feature extractor in Fig.~\ref{fig2}) with trainable parameters $\theta_g$,
\begin{small}
\begin{equation}\label{eq2}
c_k^\mathcal{S}=\frac{1}{N_s^k}\sum\limits_{(x_i^s,y_i^s)\in{D_s^k}}G(x_i^s),
\end{equation}
\end{small}
where $D_s^k$ denotes the set of samples labeled with class $k$ in the source domain and $N_s^k$ is the number of corresponding samples. Then, a set of prototypes $\{c_k^\mathcal{S}\}_{k=1}^C$ are obtained. The embedded target samples are supposed to gather around the source prototypes in the latent feature space. Thus, we use a similarity measurement $\psi$ to cluster $j$-th unlabeled target sample, $x_j^t$, to the corresponding source prototypes, where $\psi$ is computed as follows,
\begin{small}
\begin{equation}\label{eq3}
\psi(x_j^t)=CS(G(x_j^t),c_k^\mathcal{S}), k=\{1,2,...,C\}
\end{equation}
\end{small}
where $CS(.,.)$ denotes the cosine similarity function between two vectors. $x_j^t$ is added into the target domain of the class $D_t^{k'}$ with a pseudo-label $\hat{y}_j^t=k'$ where $k' = arg \max\limits_{k}\; \psi_k(x_j^t)$. 

Then, the unlabeled target samples $D_t$ are partitioned into $C$ classes (\emph{i.e.} $D_t=\{D_t^k\}_{k=1}^C=\{x_j^t,\hat{y}_j^t\}_{j=1}^{n_t}$) and each sample is scored by its similarity.
To obtain the ``easy'' samples, we constrain that the similarity scores should above a certain threshold $\tau$. During the training process, the values of the similarity $\psi$ increase continuously because the source samples and the target samples become closer to each other in the hidden space as training proceeds. ``Hard'' samples in the earlier stages may be selected as ``easy'' in the later stages. However, the constant threshold will turn too much ``hard'' samples into ``easy'' samples in each step. 
To control the growth rate of the ``easy'' samples, we gradually adjust the threshold step by step as follows,
\begin{small}
\begin{equation}\label{eq4}
\tau=\frac{1}{1+e^{-\mu{\cdot}(m+1)}}-0.01,
\end{equation}
where $\mu$ is a constant and $m$ ($m=\{0,1,2,...\}$) denotes the training step. Therefore, the sample selection function is formulated as follows,
\begin{equation}\label{eq5}
{\forall{x_j}\in{D_t^k}|}_{k=1}^{C}, w_j = \left\{
                            \begin{array}{lcl}
                            1 &\text{if} &\psi \ge \tau \\
                            0 &\text{if} &\psi < \tau
                            \end{array}
                            \right.
\end{equation}
\end{small}
where $w_j=1$ indicates $x_j$ to be selected; otherwise, $w_j=0$ indicates $x_j$ not to be selected. Finally, we obtain a selected pseudo-labeled target domain $\hat{D}_t=\{x_j^t,\hat{y}_j^t\}_{j=1}^{\hat{n}_t}$, where $\hat{n}_t$ denotes the number of selected samples.

\subsection{Adaptive Prototype Alignment}
\label{sec:APA}
In this section, we introduce the proposed APA, which considers the pairwise semantic similarity across domains to explicitly alleviate the negative influence of those false-easy samples and enforce the cross-domain category consistency. 
It can be implemented by aligning the prototype of source and selected target samples for each category. 
We measure the distance between two prototypes as follows,
\begin{small}
\begin{equation}\label{eq6}
d(c_k^\mathcal{S},c_k^\mathcal{T})=\left\Vert{c_k^\mathcal{S}-c_k^\mathcal{T}}\right\Vert^2,
\end{equation}
\end{small}
where $c_k^\mathcal{S}$ and $c_k^\mathcal{T}$ represent the source and target prototypes, respectively. We opt for the squared Euclidean distance as the distance measure function. The justification is that the cluster mean yields optimal cluster representatives when a Bregman divergence (\emph{e.g.} squared Euclidean distance and Mahalanobis distance) is used \cite{snell2017prototypical}. An optional approach for prototype alignment is to compute and align the local prototypes based on the mini-batch sampled from $D_s$ and $\hat{D}_t$ at each iteration. However, this approach is in a position of weakness because the categorical information in each mini-batch is expected to be insufficient, even one falsely labeled sample in the target mini-batch may cause huge bias between the computed prototype and true prototype. 

To overcome the aforementioned problems, we propose to adaptively align the global prototypes. The APA first computes the initial global prototypes based on the selected pseudo-labeled target samples $\hat{D}_t$ as follows,
\begin{small}
\begin{equation}\label{eq7}
c_{k(0)}^\mathcal{T}=\frac{1}{\hat{D}_t^k}\sum\limits_{(x_j^t,y_j^t)\in{\hat{D}_t^k}}G(x_j^t).
\end{equation}
\end{small}
In each iteration, we compute a set of local prototypes $\{c_k^t\}_{k=1}^C$ using the mini-batch samples. The accumulated prototypes are computed as the average of all previous local prototypes in each iteration,
\begin{small}
\begin{equation}\label{eq8}
\overline{c}_{k(I)}^t=\frac{1}{I}\sum\limits_{i=1}^{I}{c_k^t}_{(i)},
\end{equation}
\end{small}
where $I$ denotes the iteration times in the current training step. Then, the new $c_k^T$ are updated as follows,
\begin{small}
\begin{equation}\label{eq9}
\begin{split}
\rho_t&=CS(\overline{c}_{k(I)}^t,c_{k(I-1)}^\mathcal{T}),\\
c_{k(I)}^\mathcal{T}&\leftarrow{{\rho_t}^2{\overline{c}_{k(I)}^t}+(1-{\rho_t}^2)c_{k(I-1)}^\mathcal{T}},
\end{split}
\end{equation}
\end{small}
where $CS(.,.)$ is the cosine distance which was defined in Eq.~\eqref{eq3} and $\rho$ is the trade-off parameters. let $c_{k(I)}^\mathcal{S}$ be analogously updated for the source domain. To this end, the APA loss is formulated as follows,
\begin{small}
\begin{equation}\label{eq19}
\mathcal{L}_{apa}(\theta_g)=\sum\limits_{k=1}^{C}d(c_{k(I)}^\mathcal{S},c_{k(I)}^\mathcal{T}).
\end{equation}
\end{small}
The motivations of APA is intuitive: 
1) the accumulated prototypes are introduced to estimate the accumulated shift caused by the falsely labeled samples, and then we can use their similarity with the previous global prototypes to decide the new global prototypes $c_k^\mathcal{T}$; and 2) we statistically align the cross-domain category distributions which can alleviate the error accumulation of the pseudo-labels.
\begin{algorithm}[!t]
\caption{Progressive Feature Alignment Network, $m=\{0,1,...\}$ denotes the training step, $I$ denotes the iteration times, $B_s$ and $B_t$ denote the mini-batch training sets.}\label{algorithm1}
  \begin{algorithmic}[1]
  \REQUIRE labeled source samples $D_s=\{(x_i^s,y_i^s)\}_{i=1}^{n_s}$, unlabeled target samples $D_t=\{x_j^t\}_{j=1}^{n_t}$
  \ENSURE $\theta_g,\theta_d,\theta_{f}$
  \STATE $m=0$
  \STATE \textbf{\emph{Stage-1:}}
  \STATE Initialize $G$ and $F$ using $D_s$, output: $model_0$
  \STATE \textbf{\emph{Stage-2:}}
  \WHILE {not converge}
  \STATE $m=m+1$
  \STATE Run the EHTS based on $model_{m-1}$, output: $\hat{D}_t$ 
  \STATE Calculate the initial global prototypes $c_{k(0)}^\mathcal{S}$ and $c_{k(0)}^\mathcal{T}$ using $D_s$ and $\hat{D}_t$ based on $model_{m-1}$
  \FOR {$I=1$ \textbf{to} $max\_iter$}
  \STATE Derive $B_s$ and $B_t$ sampled from $D_s$ and $\hat{D}_t$
  \STATE Calculate local prototypes $c_{k(I)}^s$ and $c_{k(I)}^t$
  \STATE Update: $c_{k(I)}^\mathcal{S}$, $c_{k(I)}^\mathcal{T}$ by using Eq. \ref{eq8} and Eq. \ref{eq9}
  \STATE Train $model_m$ fine-tuned from $model_{m-1}$ using $B_s$ and $B_t$ by optimizing \ref{eq11}, output: $model_m$
  \ENDFOR
  \STATE $\hat{D}_t=\emptyset$
  \ENDWHILE
  \end{algorithmic}
\end{algorithm}
\subsection{Training Losses}
In this work, we empirically found that a good adaptor needs a \emph{non-saturated} source classifier. This empirical result is supported by the theoretical analysis described in Section~\ref{analysis}. The justification is that the adaptation model is biased towards minimizing the source classification loss, which usually converges rapidly since the available of the source true labels. However, this bias may lead the overfitting to the source samples and resulting in a limited target performance. Inspired by~\cite{hinton2015distilling}, we propose to add a high temperature variate $T$ ($T>1$) to the source classifier (as depicted in Fig.~\ref{fig2}). By that means we can retard the convergence speed of the source classification loss and effectively guides the adaptor to a better adaptation performance. We achieve this behavior via the following softmax function,
\begin{small}
\begin{equation}\label{softmax}
q_i=\frac{exp({z_i}/T)}{\sum_j exp({z_j}/T)},
\end{equation}
\end{small}
where $q_i$ denotes the class probabilities for a source samples and $z$ is the logit that produced by source classifier. Using a higher value for $T$ produces a softer output and naturally retards the convergence speed.

Adversarial learning has been successfully introduced to UDA by extracting domain-invariant features to achieve domain alignment~\cite{ganin2015unsupervised}. However, the learned representations can not ensure category alignment, which is the main source of performance reduction.  
Therefore, our work simultaneously considers domain-level and category-level alignment. 
In our PFAN, the input $x$ is first embedded by $G$ to a $D$-dimensional feature vector $\textbf{f}\in{R^D}$, i.e. $\textbf{f}=G(x;\theta_g)$. In order to make $\textbf{f}$ domain-invariant, the parameters $\theta_g$ of feature extractor $G$ are expected to be optimized by maximizing the loss of the domain discriminator $D$, while the parameters $\theta_d$ of domain discriminator $D$ are trained by minimizing the loss of the domain discriminator, the discriminator is optimized following a standard classification loss:
\begin{small}
\begin{equation}\label{eq10}
\begin{split}
\mathcal{L}_d(\theta_g,\theta_d)&=E_{x\sim{D_s}}[logD(G(x))]\\
&+E_{x\sim{\hat{D}_t}}[logD(1-G(x))],
\end{split}
\end{equation}
\end{small}

In addition, we also need to simultaneously minimize the loss of the label predictor $F$ for the labeled source samples and the APA loss. Formally, our ultimate goal is to optimize the following minimax objective:
\begin{small}
\begin{equation}\label{eq11}
\begin{split}
\min\limits_{\theta_g,\theta_{f}}\max\limits_{\theta_d}\;\sum\limits_{i=1}^{n_s}&\mathcal{L}_c(F(G(x_i^s;\theta_g);\theta_{f}),y_i^s)\\
+&\lambda\mathcal{L}_d(\theta_g,\theta_d)+\gamma\mathcal{L}_{apa}(\theta_g)
\end{split}
\end{equation}
\end{small}
where $\mathcal{L}_c$ is the standard cross-entropy loss, $\lambda$ and $\gamma$ are weights that control the interaction among the source classification loss,
the domain confusion loss and the APA loss. The pseudo-code of training PFAN is shown in Algorithm~\ref{algorithm1}, the EHTS and APA work alternatively and iteratively.
\subsection{Theoretical Analysis}
\label{analysis}
In this section, we theoretically show that our approach improves the boundary of the expected error on the target samples, making use of the theory of domain adaptation \cite{ben2010theory}. Formally, let $\mathcal{H}$ be the hypothesis class and given two domains $\mathcal{S}$ and $\mathcal{T}$,
the probabilistic bound of the error of hypothesis $h$ on the target domain is defined as,
\begin{equation}
\forall{h}\in{\mathcal{H}},R_{\mathcal{T}}(h)\leq{R_{\mathcal{S}}(h)}+\frac{1}{2}d_{\mathcal{H}\Delta\mathcal{H}}(\mathcal{S},\mathcal{T})+C\label{eq:bound}
\end{equation}
\noindent
where the expected error on the target samples, $R_{\mathcal{T}}(h)$, are bounded by three terms: (1) the expected error on the source domain, $R_{\mathcal{S}}(h)$; (2) $d_{\mathcal{H}\Delta\mathcal{H}}(\mathcal{S},\mathcal{T})$ is the domain divergence measured by a discrepancy distance between two distributions $\mathcal{S}$ and $\mathcal{T}$ \emph{w.r.t.} a hypothesis set $\mathcal{H}$; (3) the shared error of the ideal joint hypothesis, $C$.

In Inequality~\eqref{eq:bound}, $R_{\mathcal{S}}(h)$ is expected to be small and prone to be optimized by a deep network since we have source labels. On the other hand, prior efforts \cite{ganin2015unsupervised} seeks to minimize $d_{\mathcal{H}\Delta\mathcal{H}}(\mathcal{S},\mathcal{T})$ by the domain classifier-based adversarial learning. However, A small $d_{\mathcal{H}\Delta\mathcal{H}}(\mathcal{S},\mathcal{T})$ and a small $R_{\mathcal{S}}(h)$ do not guarantee small $R_{\mathcal{T}}(h)$. It is possible that $C$ tends to be large when the cross-domain category alignment is not be explicitly enforced (\emph{i.e.} the marginal distribution is well aligned, but the class conditional distribution is not guaranteed). Therefore, $C$ needs to be bounded as well.
Unfortunately, we cannot directly measure $C$ due to the absence of target true labels. Thus, we resort to the pseudo-labels to give the approximate evaluation and minimization.
\begin{figure}[!t]
\centering
\setlength{\belowcaptionskip}{-0.5cm}
\includegraphics[width=7cm]{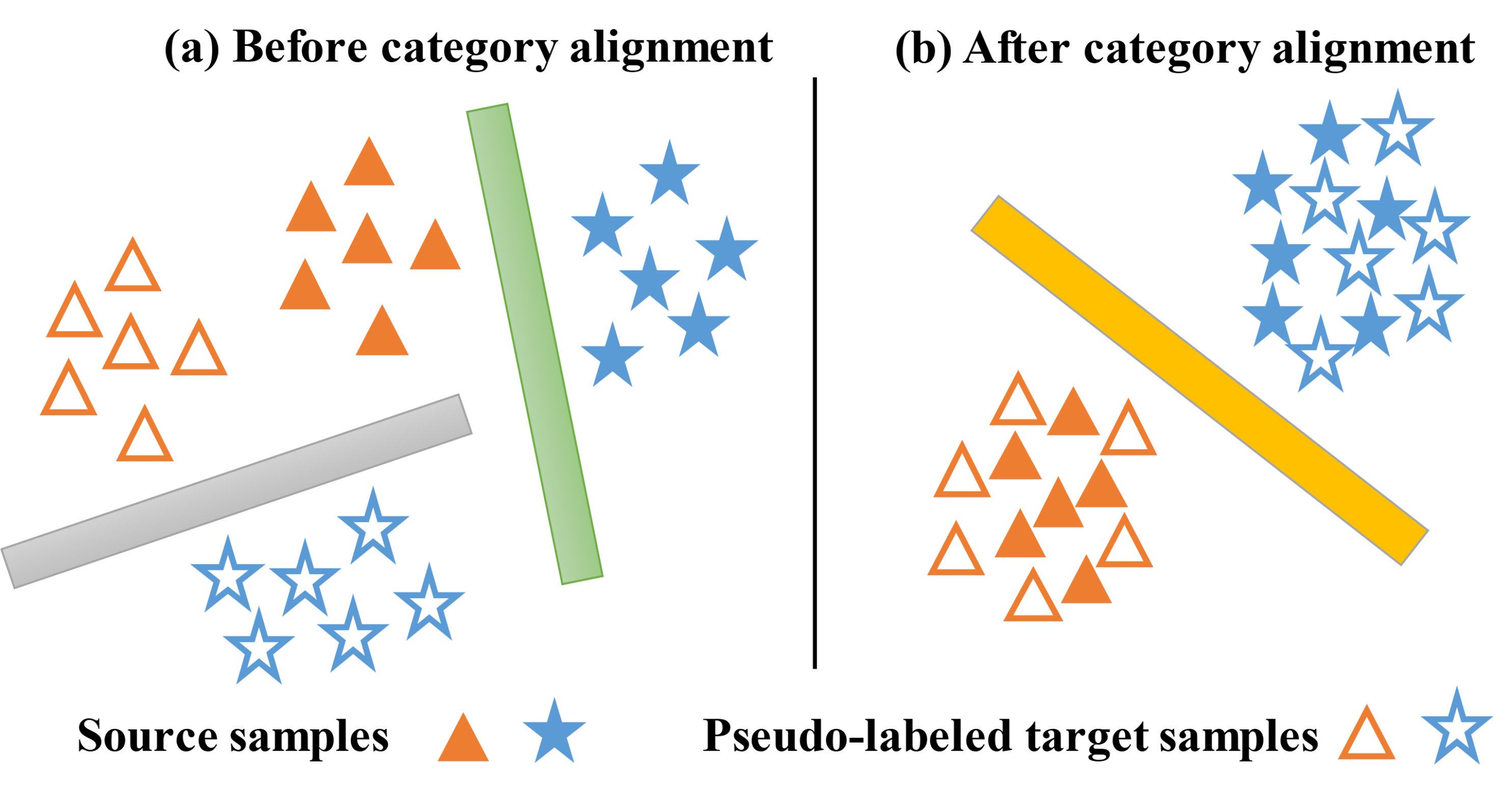}
\caption{(a) Before category alignment: there exists an optimality gap. (b) After category alignment: the optimality gap does not exist any more.
}\label{fig:analysis}
\end{figure}
\begin{definition}
If $R_{\mathcal{T}'}(\cdot)$ denotes the expected risk on the selected pseudo-labeled target set $\hat{D}_t$, the ideal joint hypothesis is the hypothesis which minimizes the combined error
\begin{equation*}
h^*=\arg\min\limits_{{h}\in\mathcal{H}}\;R_{\mathcal{S}}(h,f_{\mathcal{S}})+R_{\mathcal{T}'}(h,f_{\mathcal{T}}),
\end{equation*}
and the combined error of the ideal hypothesis is
\begin{equation}\label{eq14}
C=R_{\mathcal{S}}(h^*,f_{\mathcal{S}})+R_{\mathcal{T}'}(h^*,f_{\mathcal{T}}),
\end{equation}
where $f_{\mathcal{S}}$ and $f_{\mathcal{T}}$ are the labeling functions for the source and target domains, respectively.
\end{definition}
To bound the combined error of the ideal hypothesis, the following inequality holds:
\begin{theorem}
\label{eq13}
Let $f_{\mathcal{\hat{T}}}$ be the pseudo-labeling function. Given $R_{\mathcal{T'}}(f_{\mathcal{S}},f_{\mathcal{\hat{T}}})$ and $R_{\mathcal{T'}}(f_{\mathcal{T}},f_{\mathcal{\hat{T}}})$ as the minimum shared error and the degree to which the target samples are falsely labeled on $\hat{D}_t$, respectively. We have
\begin{small}
\begin{equation}
C\leq \min\limits_{{h}\in\mathcal{H}} R_{\mathcal{S}}(h,f_{\mathcal{S}})+R_{\mathcal{T'}}(h,f_{\mathcal{\hat{T}}})+2R_{\mathcal{T'}}(f_{\mathcal{S}},f_{\mathcal{\hat{T}}})+R_{\mathcal{T'}}(f_{\mathcal{T}},f_{\mathcal{\hat{T}}}).
\end{equation}
\end{small}
\noindent
\end{theorem}
We show the derivation of Theorem \ref{eq13} in the Supplementary Material. 
It is easy to respectively find a suitable $h$ in $\mathcal{H}$ to approximate the $f_{\mathcal{S}}$ and $f_{\mathcal{\hat{T}}}$ since we have the source labels and target pseudo-labels. However, we assume that when the category alignment has not been achieved, there exists an optimality gap between $f_{\mathcal{S}}$ and $f_{\mathcal{\hat{T}}}$ (Fig.~\ref{fig:analysis}(a)).
%
While most existing methods do not consider such phenomenon and directly minimizing $R_{\mathcal{S}}(h,f_{\mathcal{S}})$, which leads the overfitting to source samples. 
\begin{remark}[Minimizing $R_{\mathcal{S}}(h,f_{\mathcal{S}})+R_{\mathcal{T'}}(h,f_{\mathcal{\hat{T}}})$]
The proposed softmax function with a temperature variate alleviates the overfitting to source samples (i.e. enforcing a non-saturated source classifier) by retarding the convergence speed of $R_{\mathcal{S}}(h,f_{\mathcal{S}})$. This guides the adaptation model to a better target performance, i.e., a smaller $R_{\mathcal{S}}(h,f_{\mathcal{S}})+R_{\mathcal{T'}}(h,f_{\mathcal{\hat{T}}})$. Note that when the cross-domain category distributions is well aligned, the aforementioned optimality gap is removed (Fig.~\ref{fig:analysis}(b)).
\end{remark}

Recall that the labeling function $f$ can be decomposed into the feature extractor $G$ and label classifier $F$. By considering the 0-1 loss function $\sigma$ for $R_{\mathcal{T'}}$, we have
\begin{footnotesize}
\begin{equation}\label{eq18}
\begin{split}
&R_{\mathcal{T'}}(f_{\mathcal{S}},f_{\mathcal{\hat{T}}})=E_{{x}\sim{\mathcal{T'}}}[\sigma(F_{\mathcal{S}}(G(x)),F_{\mathcal{\hat{T}}}(G(x)))]\\
&= E_{{x}\sim{\mathcal{T'}}}[|\sigma(F_{\mathcal{S}}(G(x)),y_1)-\sigma(F_{\mathcal{\hat{T}}}(G(x)),y_2)|]
\end{split}
\end{equation}
\end{footnotesize}
where
\begin{footnotesize}
\begin{equation}\label{eq19}
|\sigma(F_{\mathcal{S}}(G(x)),y_1)-\sigma(F_{\mathcal{\hat{T}}}(G(x)),y_2)|\\
=\left\{
                            \begin{array}{lcl}
                            1 &\text{if} &{y_1}\neq{y_2} \\
                            0 &\text{if} &{y_1}={y_2}
                            \end{array}
                            \right.
\end{equation}
\end{footnotesize}
\begin{remark}[Minimizing shared error]
The proposed approach aims to progressively align feature in category-level, i.e., it aligns the $k$th class in source domain $D_s^k$ with the same pseudo-labeled target class $\hat{D}_t^k$.
When the categories are aligned, it is safe to assume that $y_1=y_2$. Thus, $R_{\mathcal{T'}}(f_{\mathcal{S}},f_{\mathcal{\hat{T}}})$ is expected to be minimized.
\end{remark}

\begin{remark}[Minimizing the degree to which the target samples are falsely labeled on $\hat{D}_t$]
The proposed EHTS aims to select reliable pseudo-labeled samples in the target domain which minimizes $R_{\mathcal{T'}}(f_{\mathcal{T}},f_{\mathcal{\hat{T}}})$.
\end{remark}

\vspace{-0.4cm}
\section{Experiments}
\subsection{Datasets and Baselines}
%
%
\textbf{Office-31}~\cite{saenko2010adapting} is a popular benchmark for evaluation on domain adaptation. It contains $4110$ images of $31$ categories in total, which are collected from three domains, including Amazon ($\textbf{A}$) comprising 2817 images downloaded from online merchants, Webcam ($\textbf{W}$) involving 795 low resolution images acquired from webcams, and DSLR ($\textbf{D}$) containing 498 high resolution images of digital SLRs. We try all 6 combinations of two domains for evaluation.

\textbf{ImageCLEF-DA}~\cite{caputo2014imageclef} originally used for the ImageCLEF 2014 domain adaptation challenge consists of twelve common classes from three domains: ImageNet ILSVRC 2012 (\textbf{I}), Pascal VOC 2012 (\textbf{P}), and Caltech-256 (\textbf{C}). Each doamin has 600 images in total and contains 50 images per class.  We test 6 tasks by using all domain combinations.

\textbf{MNIST}~\cite{lecun1998gradient}, \textbf{SVHN}~\cite{netzer2011reading} and \textbf{USPS}~\cite{denker1989neural} contain digital images of $10$ classes. In particular, the images in MNIST and SVHN are grey, and are of size $28\times28$ and $16\times16$, respectively; USPS consists of color images of size $32\times32$, and there are often more than one digit in one image. Following previous works, we consider the three transfer tasks: MNIST$\rightarrow$SVHN, SVHN$\rightarrow$MNIST and MNIST$\rightarrow$USPS.
\vspace{-0.1cm}
\subsection{Implementation Details}
Joining previous practices, we instantiate our backbone by AlexNet that has been pre-trained on ImageNet for Office-31 and ImageCLEF-DA, and employ the CNN architecture by~\cite{tzeng2017adversarial} for the digital datasets. 
As suggested by~\cite{long2017deep}, we fine-tune the feature extractor $G$ upon the backbone and train the predictor $F$ from the scratch via back propagation.
We utlize stochastic gradient descent (SGD) for the training with a momentum of 0.9 and a annealing learning rate (lr) given by $lr_p=\frac{lr_0}{(1+\alpha{p})^{\beta}}$, where $p$ is increased linearly from 0 to 1 as the training proceeds, $lr_0=0.01$, $\alpha=10$, and $\beta=0.75$.
In order to suppress noisy signal especially for the initial training steps, we use the similar schedule method as \cite{ganin2015unsupervised} to adaptively change the values of $\lambda$ and $\gamma$ in Eq.~\eqref{eq11} by computing $\lambda=\gamma=\frac{2}{1+exp(-\delta{p})}-1$ with $\delta=10$. We set $T=1.8$ in Eq.~\eqref{softmax} and $\mu=0.8$ in Eq.~\eqref{eq4} for all experiments.
The batch size is selected as 128. The means and standard derivations of all results are obtained over 5 random runs. All experiments are implemented by the Caffe framework.
\begin{center}
\begin{table*}[thb]
\small
\caption{AlexNet-based approaches on Office-31 (\%)}\label{lab1}
\centering
\setlength{\tabcolsep}{-2mm}{
\begin{tabular*}{\hsize}{@{}@{\extracolsep{\fill}}cccccccc@{}}
\toprule
Method & A $\rightarrow$ W & D $\rightarrow$ W & W $\rightarrow$ D & A $\rightarrow$ D & D $\rightarrow$ A & W $\rightarrow$ A & Avg\\
\hline
AlexNet \cite{krizhevsky2012imagenet} & 61.5$\pm$0.5 & 95.1$\pm$0.3 & 99.0$\pm$0.2 & 64.4$\pm$0.5 & 48.8$\pm$0.3 & 47.0$\pm$0.4 & 69.3\\
DDC \cite{tzeng2014deep}  & 61.8$\pm$0.4 & 95.0$\pm$0.5 & 98.5$\pm$0.4 & 64.4$\pm$0.3 & 52.1$\pm$0.6 & 52.2$\pm$0.4 & 70.6\\
DAN \cite{long2015learning}  & 68.5$\pm$0.4 & 96.0$\pm$0.3 & 99.0$\pm$0.2 & 67.0$\pm$0.4 & 54.0$\pm$0.4 & 53.1$\pm$0.3 & 72.9\\
RTN \cite{long2016unsupervised}  & 73.3$\pm$0.3 & 96.8$\pm$0.2 & 99.6$\pm$0.1 & 71.0$\pm$0.2 & 50.5$\pm$0.3 & 51.0$\pm$0.1 & 73.7\\
RevGrad \cite{ganin2015unsupervised}  & 73.0$\pm$0.5 & 96.4$\pm$0.3 & 99.2$\pm$0.3 & 72.3$\pm$0.3 & 53.4$\pm$0.4 & 51.2$\pm$0.5 & 74.3\\
JAN \cite{long2017deep}  & 74.9$\pm$0.3 & 96.6$\pm$0.2 & 99.5$\pm$0.2 & 71.8$\pm$0.2 & 58.3$\pm$0.3 & 55.0$\pm$0.4 & 76.0\\
MADA \cite{pei2018multi}  &78.5$\pm$0.2 & \textbf{99.8}$\pm$0.1 & \textbf{100.0}$\pm$.0 & 74.1$\pm$0.1 & 56.0$\pm$0.2 & 54.5$\pm$0.3 & 77.1\\
MSTN \cite{xie2018learning}  &80.5$\pm$0.4 & 96.9$\pm$0.1 & 99.9$\pm$0.1 & 74.5$\pm$0.4 & 62.5$\pm$0.4 & 60.0$\pm$0.6 & 79.1\\
\hline
PFAN & \textbf{83.0}$\pm$0.3 & 99.0$\pm$0.2 & 99.9$\pm$0.1 & \textbf{76.3}$\pm$0.3 & \textbf{63.3}$\pm$0.3 & \textbf{60.8}$\pm$0.5 & \textbf{80.4}\\
\bottomrule
\vspace{-0.55cm}
\end{tabular*}}
\end{table*}
\end{center}
\begin{center}
\begin{table*}[htb]
\small
\caption{AlexNet-based approaches on ImageCLEF-DA (\%)}\label{lab2}
\centering
\setlength{\belowcaptionskip}{-0.5cm}
\setlength{\tabcolsep}{-2mm}{
\begin{tabular*}{\hsize}{@{}@{\extracolsep{\fill}}cccccccc@{}}
\toprule
Method & I $\rightarrow$ P & P $\rightarrow$ I & I $\rightarrow$ C & C $\rightarrow$ I & C $\rightarrow$ P & P $\rightarrow$ C & Avg\\
\hline
AlexNet \cite{krizhevsky2012imagenet} & 66.2$\pm$0.2 & 70.0$\pm$0.2 & 84.3$\pm$0.2 & 71.3$\pm$0.4 & 59.3$\pm$0.5 & 84.5$\pm$0.3 & 73.9\\
DAN \cite{long2015learning}  & 67.3$\pm$0.2 & 80.5$\pm$0.3 & 87.7$\pm$0.3 & 76.0$\pm$0.3 & 61.6$\pm$0.3 & 88.4$\pm$0.2 & 76.9\\
RevGrad \cite{ganin2015unsupervised}  & 66.5$\pm$0.5 & 81.8$\pm$0.4 & 89.0$\pm$0.5 & 79.8$\pm$0.5 & 63.5$\pm$0.4 & 88.7$\pm$0.4 & 78.2\\
JAN \cite{long2017deep}  & 67.2$\pm$0.5 & 82.8$\pm$0.4 & 91.3$\pm$0.5 & 80.0$\pm$0.5 & 63.5$\pm$0.4 & 91.0$\pm$0.4 & 79.3\\
MADA \cite{pei2018multi}  &68.3$\pm$0.3 & 83.0$\pm$0.1 & 91.0$\pm$0.2 & 80.7$\pm$0.2 & 63.8$\pm$0.2 & \textbf{92.2}$\pm$0.3 & 79.8\\
MSTN \cite{xie2018learning}  &67.3$\pm$0.3 & 82.8$\pm$0.2 & 91.5$\pm$0.1 & 81.7$\pm$0.3 & 65.3$\pm$0.2 & 91.2$\pm$0.2 & 80.0\\
\hline
PFAN & \textbf{68.5}$\pm$0.5 & \textbf{84.4}$\pm$0.4 & \textbf{92.2}$\pm$0.6 & \textbf{82.3}$\pm$0.4 & \textbf{66.3}$\pm$0.3 & 91.7$\pm$0.2 & \textbf{80.9}\\
\bottomrule
\vspace{-0.75cm}
\end{tabular*}}
\end{table*}
\end{center}
\vspace{-1.9cm}
\subsection{Comparisons with State-of-the-Arts}
\paragraph{State-of-the-arts.} We compare our approach with various state-of-the-art UDA methods, including \textbf{AlexNet}~\cite{krizhevsky2012imagenet}, Deep Domain Confusion (\textbf{DDC})~\cite{tzeng2014deep}, 
Deep Adaptation Network (\textbf{DAN})~\cite{long2015learning}, Residual Transfer Network (\textbf{RTN})~\cite{long2016unsupervised} , Reverse Gradient (\textbf{RevGrad})~\cite{ganin2015unsupervised}, Adversarial Discriminative Domain Adaptation  (\textbf{ADDA})~\cite{tzeng2017adversarial}, 
Joint Adaptation Networks (\textbf{JAN})~\cite{long2017deep}, Asymmetric Tri-Training (\textbf{ATT})~\cite{saito2017asymmetric} , Multi-Adversarial Domain Adaptation (\textbf{MADA})~\cite{pei2018multi}, and Moving Semantic Transfer Network (\textbf{MSTN})~\cite{xie2018learning}. For all above methods, we summarize the results reported in their original papers. For similarity, we term our method as PFAN hereafter.

Table~\ref{lab1} displays the results on Office-31. The proposed PFAN outperforms all compared methods in general and improves the state-of-the-art result from $79.1\%$ to $80.4\%$ on average.
If we focus more on the hard transfer tasks (\emph{e.g.}  $\textbf{A}\rightarrow\textbf{W}$ and $\textbf{A}\rightarrow\textbf{D}$), PFAN substantially exhibits better transferring ability than others.
In contrast to JAN, MADA and MSTN, our PFAN additionally considers both the target intra-class variation and the non-saturated source classifier. Our better performance over them could indicate the effectiveness of these two components.
RevGrad has also taken the domain adversarial adaptation into account, but its results are still inferior to ours. The advantage of our model compared to RevGrad is that, we furhter perform EHTS and APA, which as supported by our experiments can explicitly enforce the cross-domain category alignment, hence delivering better performance.

The results of ImageCLEF-DA are reported in Table \ref{lab2}. Our approach outperforms all comparison methods on most transfer tasks, which reveals that PFAN is scalable for different datasets. 

The results of digit classification are reported in Table~\ref{lab3}. We follow the training protocol established in \cite{tzeng2017adversarial}. For adaptation between MNIST and USPS, we randomly sample 2000 images from MNIST and 1800 from USPS. For adaptation between SVHN and MNIST, we use the full training sets. For the hard transfer task MNIST$\rightarrow$SVHN, we reproduced the MSTN \cite{xie2018learning} but were unable to get it to converge, since the performance of this approach depends strongly on the accuracy of the pseudo-labeled samples which was lower on this task. In contrast, our approach significantly outperforms the suboptimal result by +4.8\%, which clearly demonstrates the effect of our approach on selecting reliable pseudo-labeled samples and alleviating the negative influence of falsely-labeled samples on the challenging scenario. For the easier tasks SVHN$\rightarrow$MNIST and MNIST$\rightarrow$USPS, our approach also shows superiority.
\begin{center}
\begin{table}[!t]
\footnotesize
\caption{Accuracy on the digit classification task. (\%)}\label{lab3}
\centering
\setlength{\belowcaptionskip}{-2cm}
\setlength{\tabcolsep}{-2mm}{
\begin{tabular*}{\hsize}{@{}@{\extracolsep{\fill}}cccc@{}}
\toprule
Source & MNIST  & SVHN & MNIST  \\
Target & SVHN  & MNIST & USPS \\
\hline
Source Only & 33.0$\pm$1.2 & 60.1$\pm$1.1 & 75.2$\pm$1.6  \\
RevGrad \cite{ganin2015unsupervised}  & 35.7 & 73.9 & 77.1$\pm$1.8  \\
ADDA \cite{tzeng2017adversarial}  & - & 76.0$\pm$1.8 & 89.4$\pm$0.2  \\
ATT \cite{saito2017asymmetric}  & 52.8 & 85.0 & -  \\
MSTN \cite{xie2018learning} & did not converge& 91.7$\pm$1.5 & 92.9$\pm$1.1 \\
\hline
PFAN & \textbf{57.6}$\pm$1.8 & \textbf{93.9}$\pm$0.8 & \textbf{95.0}$\pm$1.3 \\
\bottomrule
\vspace{-0.65cm}
\end{tabular*}}
\end{table}
\end{center}
\begin{center}
\begin{table}[!t]
\footnotesize
\caption{Ablation of PFAN on different transfer tasks. (\%)}\label{lab4}
\centering
\begin{tabular*}{\hsize}{@{}@{\extracolsep{\fill}}ccccc@{}}
\toprule
Model & A$\rightarrow$W & I$\rightarrow$P & SVHN$\rightarrow$MNIST \\
\hline
\hline
Source Only & 61.6 & 66.2 & 60.1 \\
PFAN (Random) & 77.0 & 67.0 & 87.2 \\
PFAN (Full) & 81.9 & 68.0 & 92.5 \\
PFAN (woAPA) & 76.4 & 67.1 & 82.0 \\
PFAN (woA) & 82.2 & 68.1 & 93.0 \\
PFAN (woT) & 80.6 & 67.9 & 92.1 \\
\hline
PFAN & \textbf{83.0} & \textbf{68.5} & \textbf{93.9} \\
\bottomrule
\vspace{-0.8cm}
\end{tabular*}
\end{table}
\end{center}
\begin{figure*}[!t]
\centering
\setlength{\belowcaptionskip}{-0.2cm}
\subfigure[Non-saturated source classifier]{
\label{fig31}
\includegraphics[width=1.6in]{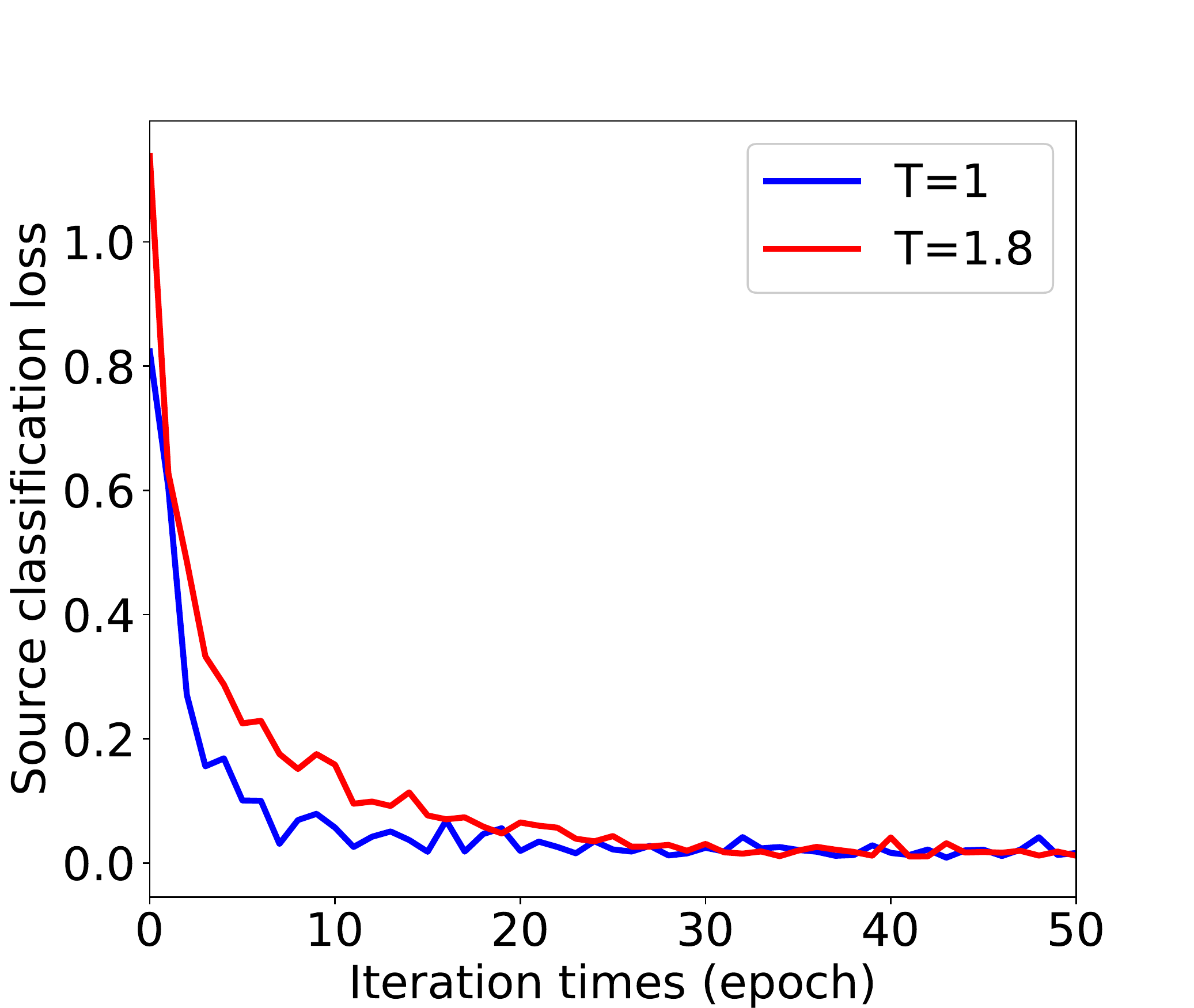}}
\subfigure[Distribution Discrepancy]{
\label{fig32}
\includegraphics[width=1.6in]{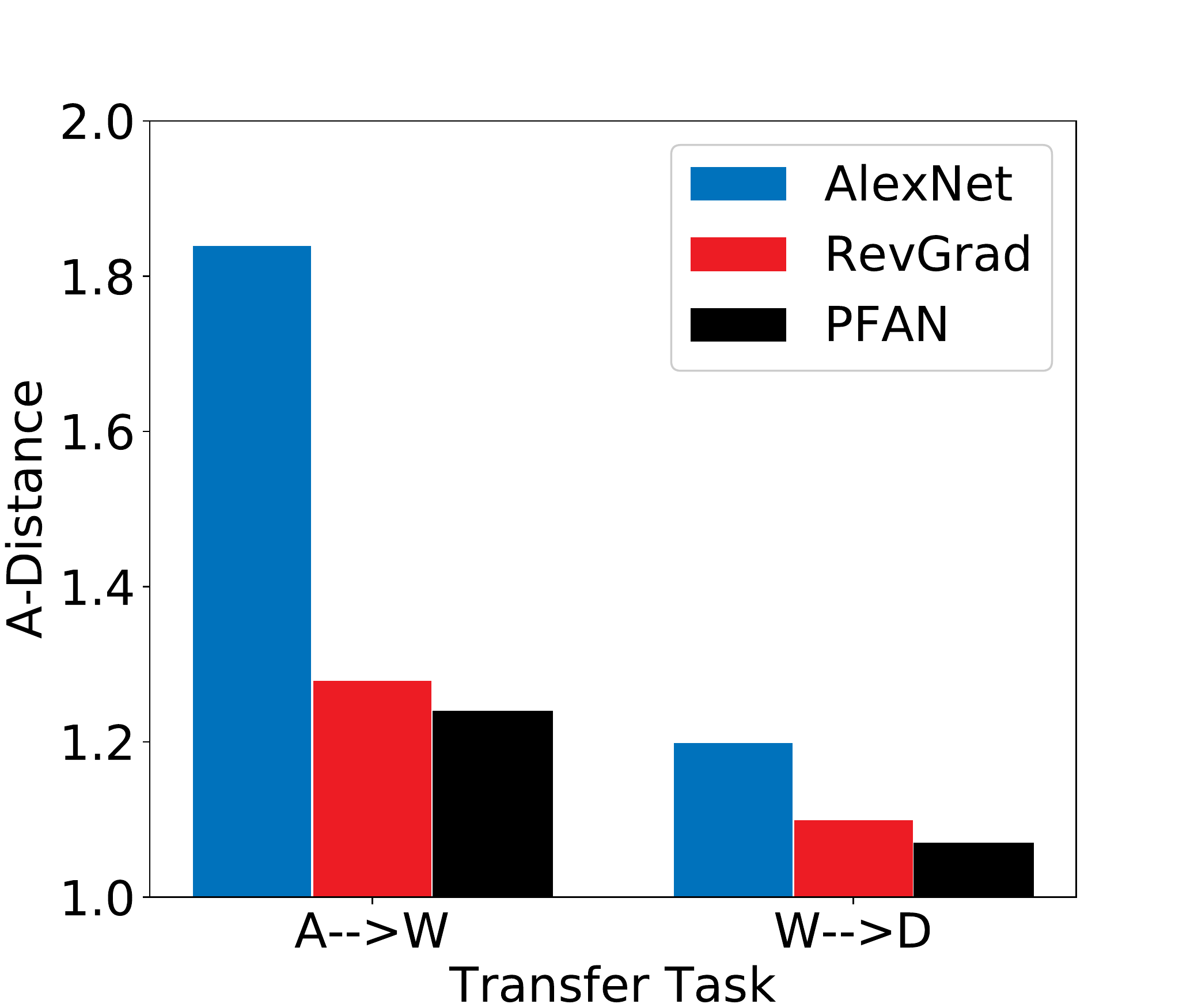}}
\subfigure[RevGrad: target=$W$]{
\label{fig33}
\includegraphics[width=1.6in]{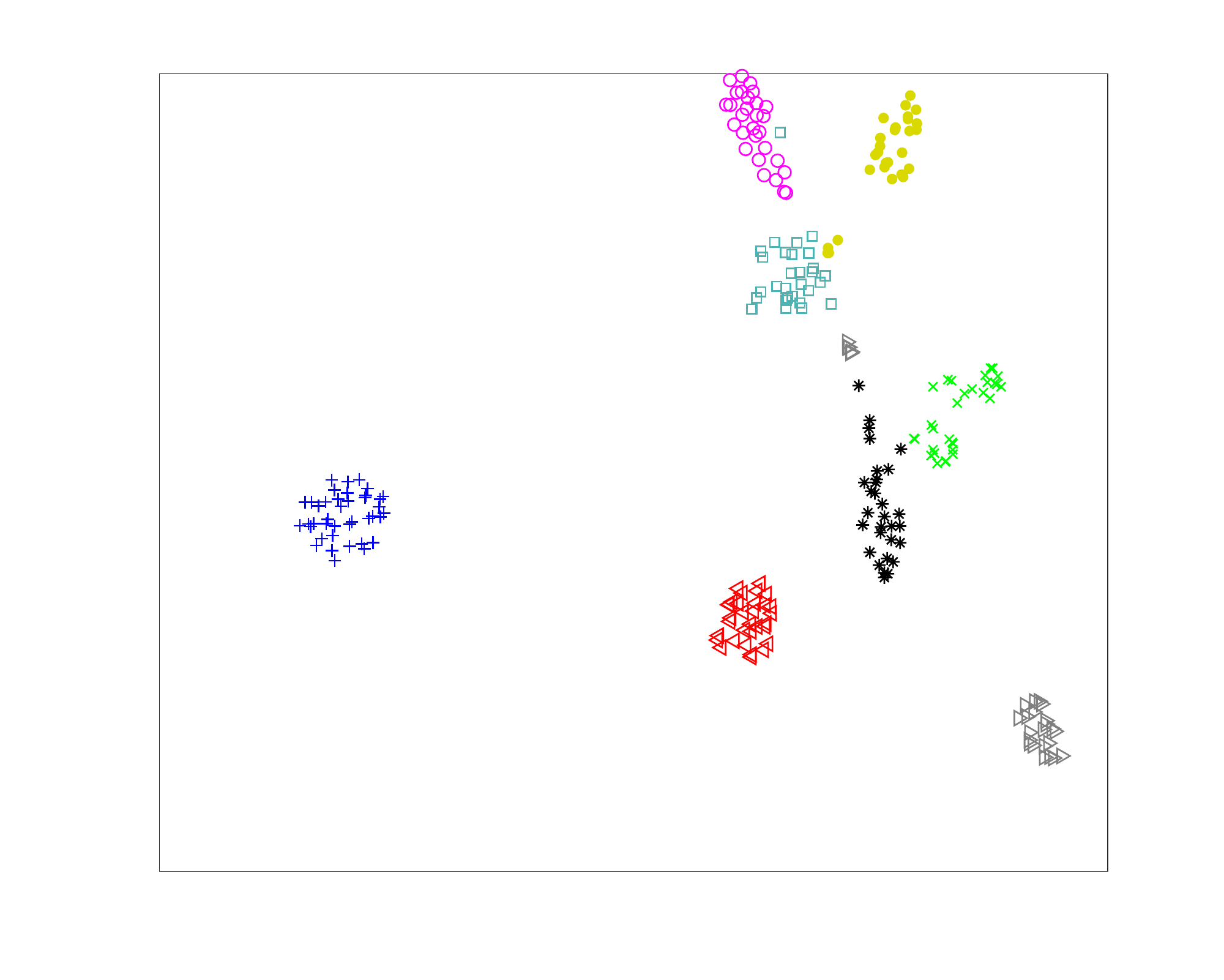}}
\subfigure[PFAN: target=$W$]{
\label{fig34}
\includegraphics[width=1.6in]{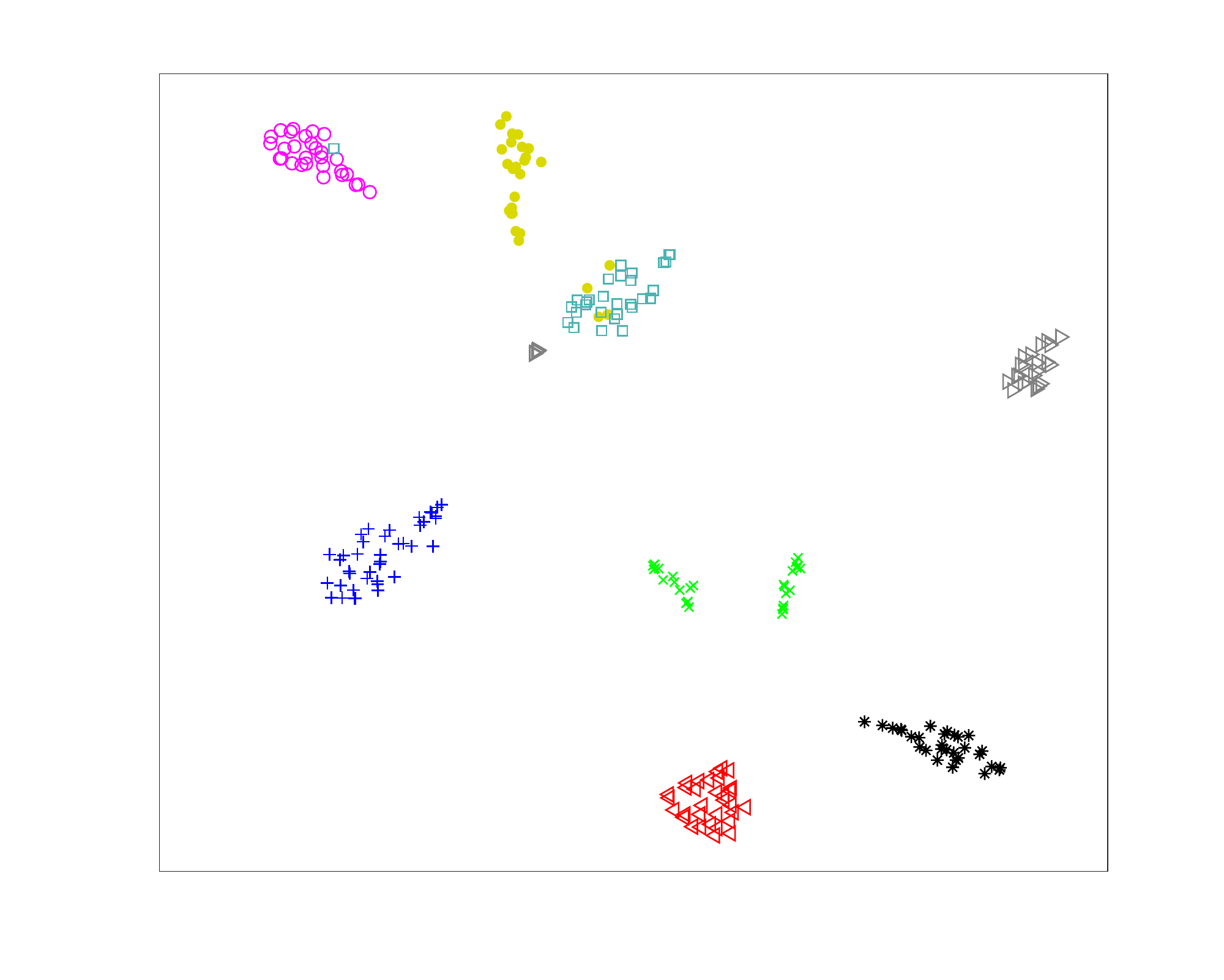}}
\caption{(a) The convergence speed of the source classification loss in different temperature setting. (b) Distribution Discrepancy. (c)-(d) The t-SNE visualization of network activations on target domain \textbf{W} generated by RevGrad and PFAN.}
\vspace{-0.5cm}
\label{fig3}
\end{figure*}
\vspace{-1.95cm}
\subsection{Further Empirical Analysis}
\paragraph{Ablation Study.}To isolate the contribution of our work, we perform ablation study by evaluating several variants of PFAN: (1) \textbf{PFAN (Random)}, which randomly selects the target samples instead of using the easy-to-hard order; (2) \textbf{PFAN (Full)}, which uses all target samples at the training period; 
(3) \textbf{PFAN (woAPA)}, which denotes training completely without the APA (i.e. $\gamma=0$ in Eq.~\eqref{eq11}); (4) \textbf{PFAN (woA)}, which denotes aligning the prototypes based on the current mini-batch without considering the global and accumulated prototypes; (5) \textbf{PFAN (woT)}, which removes the temperature from our model (i.e. $T=1$ in Eq.~\eqref{softmax}). The results are shown in Table~\ref{lab4}. We can observe that all the components are designed reasonably and when any one of these components is removed, the performance degrades. It is noteworthy that PFAN outperforms both PFAN (Random) and PFAN (Full), which reveals that the EHTS can provide more reliable and informative target samples for the cross-domain category alignment. 
\vspace{-0.5cm}

\begin{figure}[!t]
\centering
\includegraphics[width=5cm]{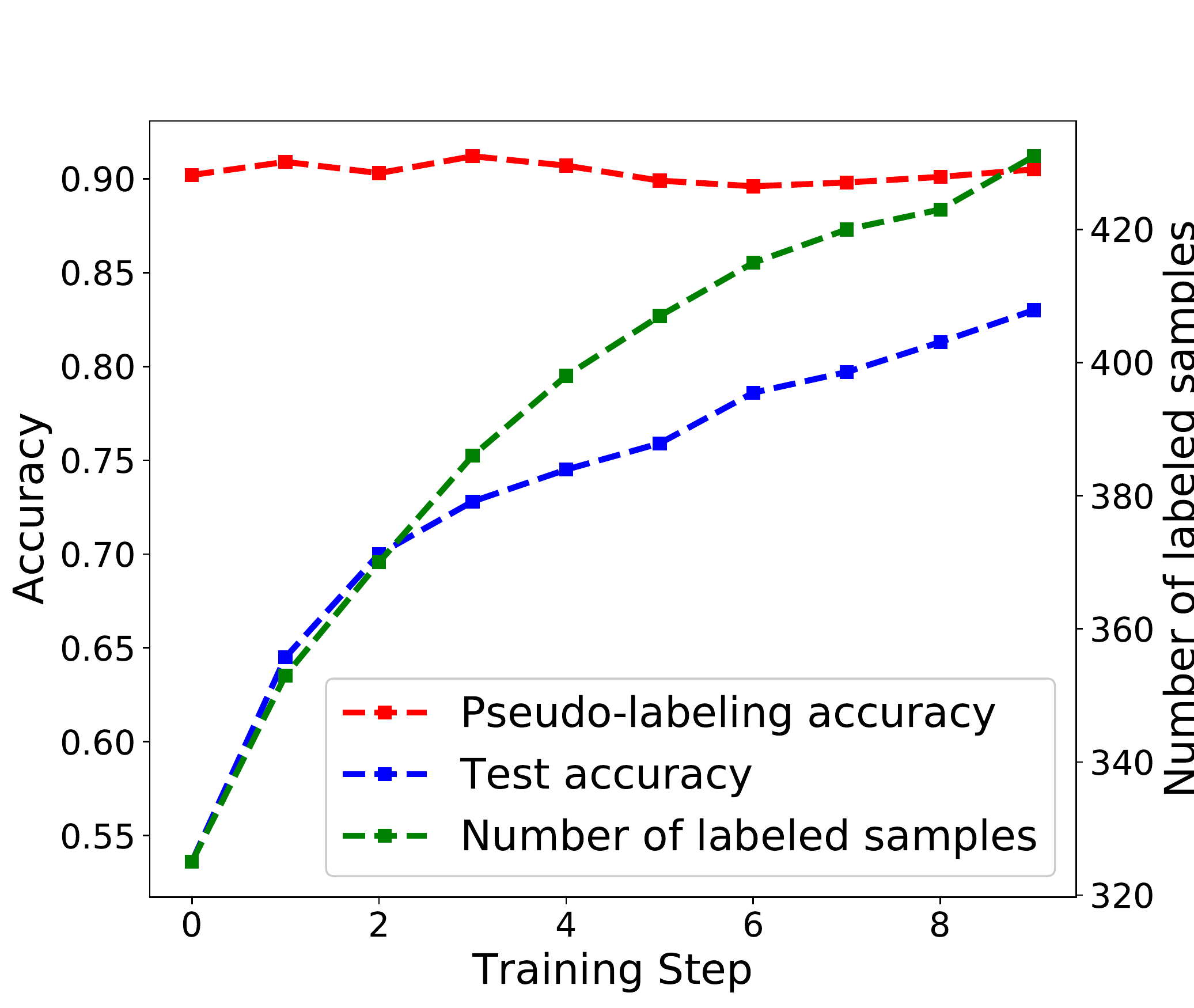}
\caption{Comparison of the pseudo-labeling accuracy and the test accuracy on transfer task A $\rightarrow$ W.
The pseudo-labeling accuracy is computed using (the number of correctly labeled samples)/(the number of labeled samples).}\label{fig4}
\vspace{-0.4cm}
\end{figure}

\paragraph{Pseudo-labeling Accuracy.} We show the relationship between the pseudo-labeling accuracy and test accuracy in Fig.~\ref{fig4}. We found that (1) the pseudo-labeling accuracy keeps higher and stable throughout as training proceeds, which thanks to the EHTS by selecting reliable pseudo-labeled samples; (2) the test accuracy increases with the increasing of labeled samples, which implies that the number of correctly and falsely labeled samples are both proportionally increasing, but our approach can explicitly alleviate the negative influence of the falsely-labeled samples. 
\vspace{-0.4cm}


\paragraph{Non-saturated source classifier.} To further verify our hypothesis about the non-saturated source classifier, we investigate the source classification loss in different temperature setting. 
The results are reported in Fig.~\ref{fig31}. The $T=1$ model converges faster than $T=1.8$ especially at the beginning of training. However, such difference gradually decreases as training proceeds.
The justification is that we use a higher $T$ to retard the convergence speed of the source classification loss (\emph{i.e.} alleviating the adaptor overfitting to the source samples), thus showing better adaptation. 
\vspace{-0.5cm}
\paragraph{Distribution Discrepancy.} The domain adaptation theory~\cite{ben2010theory} suggests that $\mathcal{A}$-distance can be used as a measure of domain discrepancy. The way of estimating empirical $\mathcal{A}$-distance was defined as $d_{\mathcal{A}}=2(1-\epsilon)$, where $\epsilon$ is the generalization error of a classifier trained to discriminate the source and target features. We utilize a kernel SVM to estimate the $\mathcal{A}$-distance. Fig.~\ref{fig32} demonstrates the $\mathcal{A}$-distance calculated with the features from AlexNet, RevGrad and PFAN on tasks $\textbf{A}\rightarrow\textbf{W}$ and $\textbf{W}\rightarrow\textbf{D}$. We can observe that our method significantly reduces the $\mathcal{A}$-distance compared with the AlexNet. However, when compared with RevGrad, PFAN shows smaller improvement with respect to $\mathcal{A}$-distance, but improves the performance by large margin, which demonstrates that a low domain divergence does not imply better performance in the target domain. This phenomenon is consistent with the analysis in Section~\ref{analysis}.
\vspace{-0.4cm}
\paragraph{Feature Visualization.} We utilize t-SNE~\cite{donahue2014decaf} to visualize the deep feature of the network activations on task $\textbf{A}\rightarrow\textbf{W}$ (randomly selected 8 classes) learned by RevGrad (the bottleneck layer) and PFAN (the bottleneck layer). As shown in Fig.~\ref{fig33}-\ref{fig34}, we can see that the RevGrad features on target domain can not be discriminated very well, some categories have been mixed up in the feature space. By contrast, PFAN can learn more discriminative representations, which jointly enlarges the inter-class dispersion and reduces the intra-class variations. 
\vspace{-0.2cm}
\section{Conclusion}
\label{conclusion}
In this paper, we proposed a novel approach called Progressive Feature Alignment Network, to take advantage of target domain intra-class variance and cross-domain category consistency for addressing UDA problems. The proposed EHTS and APA complement each other in selecting reliable pseudo-labeled samples and alleviating the bias caused by the falsely-labeled samples. The performance is further improved by retarding the convergence speed of the source classification loss. The extensive experiments reveal that our approach outperforms state-of-the-art UDA approaches on three domain adaptation datasets.
\vspace{-0.2cm}
\section{Acknowledgements}
This work was supported in part by the National Natural Science Foundation of China under Grants 61571382, 81671766, 61571005, 81671674, 61671309 and U1605252, in part by the Fundamental Research Funds for the Central Universities under Grants 20720160075 and 20720180059, in part by the CCF-Tencent open fund, and the Natural Science Foundation of Fujian Province of China (No.2017J01126).
{\small
\bibliographystyle{ieee}
\bibliography{egpaper_final}
}

\end{document}